\documentclass[twocolumn]{aastex631}
\usepackage{graphicx}   
\usepackage{lipsum}

\shorttitle{The CAMELS Multifield Dataset}
\shortauthors{Villaescusa-Navarro et al.}
\graphicspath{{./}{figures/}}

\begin{document}

\title{The CAMELS Multifield Dataset: \\ Learning the Universe's Fundamental Parameters with Artificial Intelligence}

\author[0000-0002-4816-0455]{Francisco Villaescusa-Navarro}
\affiliation{Department of Astrophysical Sciences, Princeton University, Peyton Hall, Princeton NJ 08544, USA}
\affiliation{Center for Computational Astrophysics, Flatiron Institute, 162 5th Avenue, New York, NY, 10010, USA}

\correspondingauthor{Francisco Villaescusa-Navarro}
\email{villaescusa.francisco@gmail.com}

\author{Shy Genel}
\affiliation{Center for Computational Astrophysics, Flatiron Institute, 162 5th Avenue, New York, NY, 10010, USA}
\affiliation{Columbia Astrophysics Laboratory, Columbia University, New York, NY, 10027, USA}

\author[0000-0001-5769-4945]{Daniel Angl\'es-Alc\'azar}
\affiliation{Department of Physics, University of Connecticut, 196 Auditorium Road, Storrs, CT, 06269, USA}
\affiliation{Center for Computational Astrophysics, Flatiron Institute, 162 5th Avenue, New York, NY, 10010, USA}

\author[0000-0003-2911-9163]{Leander Thiele}
\affiliation{Department of Physics, Princeton University, Princeton, NJ 08544, USA}

\author{Romeel Dave}
\affiliation{Institute for Astronomy, University of Edinburgh, Royal Observatory, Edinburgh EH9 3HJ, UK}
\affiliation{Department of Physics \& Astronomy, University of the Western Cape, Cape Town 7535, South Africa}
\affiliation{South African Astronomical Observatories, Observatory, Cape Town 7925, South Africa}

\author{Desika Narayanan}
\affiliation{Department of Astronomy, University of Florida, Gainesville, FL, USA}
\affiliation{University of Florida Informatics Institute, 432 Newell Drive, CISE Bldg E251, Gainesville, FL, USA}

\author{Andrina Nicola}
\affiliation{Department of Astrophysical Sciences, Princeton University, Peyton Hall, Princeton NJ 08544, USA}

\author[0000-0002-0701-1410]{Yin Li}
\affiliation{Center for Computational Astrophysics, Flatiron Institute, 162 5th Avenue, New York, NY, 10010, USA}
\affiliation{Center for Computational Mathematics, Flatiron Institute, 162 5th Avenue, New York, NY, 10010, USA}

\author[0000-0002-0936-4279]{Pablo Villanueva-Domingo}
\affiliation{Instituto de F\'isica Corpuscular (IFIC), CSIC-Universitat de Val\`encia, E-46980, Paterna, Spain}

\author{Benjamin Wandelt}
\affiliation{Sorbonne Universite, CNRS, UMR 7095, Institut d’Astrophysique de Paris, 98 bis boulevard Arago, 75014 Paris, France}
\affiliation{Center for Computational Astrophysics, Flatiron Institute, 162 5th Avenue, New York, NY, 10010, USA}

\author{David N. Spergel}
\affiliation{Center for Computational Astrophysics, Flatiron Institute, 162 5th Avenue, New York, NY, 10010, USA}
\affiliation{Department of Astrophysical Sciences, Princeton University, Peyton Hall, Princeton NJ 08544, USA}

\author{Rachel S. Somerville}
\affiliation{Center for Computational Astrophysics, Flatiron Institute, 162 5th Avenue, New York, NY, 10010, USA}

\author{Jose Manuel Zorrilla Matilla}
\affiliation{Department of Astrophysical Sciences, Princeton University, Peyton Hall, Princeton NJ 08544, USA}

\author{Faizan G. Mohammad}
\affiliation{Waterloo Center for Astrophysics, University of Waterloo, Waterloo, ON N2L 3G1, Canada}
\affiliation{Department of Physics and Astronomy, University of Waterloo, Waterloo, ON N2L 3G1, Canada}

\author{Sultan Hassan}
\affiliation{Center for Computational Astrophysics, Flatiron Institute, 162 5th Avenue, New York, NY, 10010, USA}
\affiliation{Department of Physics \& Astronomy, University of the Western Cape, Cape Town 7535, South Africa}

\author{Helen Shao}
\affiliation{Department of Astrophysical Sciences, Princeton University, Peyton Hall, Princeton NJ 08544, USA}

\author[0000-0002-2544-7533]{Digvijay Wadekar}
\affiliation{Center for Cosmology and Particle Physics, Department of Physics, New York University, New York, NY 10003}
\affiliation{School of Natural Sciences, Institute for Advanced Study, Princeton, NJ 08540, USA}

\author{Michael Eickenberg}
\affiliation{Center for Computational Mathematics, Flatiron Institute, 162 5th Avenue, New York, NY, 10010, USA}

\author{Kaze W.K. Wong}
\affiliation{Center for Computational Astrophysics, Flatiron Institute, 162 5th Avenue, New York, NY, 10010, USA}

\author{Gabriella Contardo}
\affiliation{Center for Computational Astrophysics, Flatiron Institute, 162 5th Avenue, New York, NY, 10010, USA}

\author{Yongseok Jo}
\affiliation{Center for Theoretical Physics, Department of Physics and Astronomy, Seoul National University, Seoul 08826, Korea}

\author{Emily Moser}
\affiliation{Department of Astronomy, Cornell University, Ithaca, NY 14853, USA}

\author{Erwin T. Lau}
\affiliation{Center for Astrophysics $\vert$ Harvard \& Smithsonian, 60 Garden St, Cambridge, MA 02138, USA}

\author{Luis Fernando Machado Poletti Valle}
\affiliation{Department of Physics, Yale University, New Haven, CT 06520, USA}

\author{Lucia A. Perez}
\affiliation{Arizona State University, School of Earth and Space Exploration, 781 Terrace Mall Tempe, AZ 85287, USA}

\author{Daisuke Nagai}
\affiliation{Department of Physics, Yale University, New Haven, CT 06520, USA}
\affiliation{Department of Astronomy, Yale University, New Haven, CT 06511, USA}

\author{Nicholas Battaglia}
\affiliation{Department of Astronomy, Cornell University, Ithaca, NY 14853, USA}

\author{Mark Vogelsberger}
\affiliation{Kavli Institute for Astrophysics and Space Research, Department of Physics, MIT, Cambridge, MA 02139, USA}

\begin{abstract}
We present the Cosmology and Astrophysics with MachinE Learning Simulations (CAMELS) Multifield Dataset, CMD, a collection of hundreds of thousands of 2D maps and 3D grids containing many different properties of cosmic gas, dark matter, and stars from more than 2,000 distinct simulated universes at several cosmic times. The 2D maps and 3D grids represent cosmic regions that span $\sim$100 million light years and have been generated from thousands of state-of-the-art hydrodynamic and gravity-only N-body simulations from the CAMELS project. Designed to train machine learning models, CMD is the largest dataset of its kind containing more than 70 Terabytes of data. In this paper we describe CMD in detail and outline a few of its applications. We focus our attention on one such task, parameter inference, formulating the problems we face as a challenge to the community. We release all data and provide further technical details at \url{https://camels-multifield-dataset.readthedocs.io}.
\end{abstract}


\section{Introduction} 
\label{sec:intro}

In the era of precision cosmology, there is great interest in developing sophisticated techniques to optimally measure cosmological parameters from astrophysical datasets.  To achieve the desired precision of next-generation experiments often requires accounting for complex astrophysical effects that can impact the properties of the luminous galaxies and gas from which cosmological parameters are inferred.  

Advances in deep learning are triggering a revolution in many different disciplines, from biology to social sciences. In cosmology, deep learning is being used to carry out many different complex tasks where traditional methods were slow, inaccurate or simply nonexistent. Examples of such tasks are speeding up simulations \citep{Siyu_2018, Renan_2020}, obtaining super-resolution simulations \citep{Doogesh_2020, Yin_2020a, Yueying_2021}, cleaning up astrophysics \citep{Pablo_2020, Lucas_2020}, painting stars and gas properties on the dark matter field \citep{Jay_2019, Jacky_2019, Zhang_2019, Noah_2020, Leander_2020, 2021MNRAS.504.4024M, Yonseok_2019,ModFenSel18,TroFer19,Harrington_2021}, changing the cosmology of a simulation \citep{Giusarma_19}, generating new data with desired properties \citep{Biwei_2020, Vanessa_2020}, and detecting anomalies \citep{Kate_2020}, among many more\footnote{See, e.g., \url{https://github.com/georgestein/ml-in-cosmology} for a comprehensive list of machine learning applications in cosmology.}.

One of the main reasons behind this success is the fact that neural networks behave as universal function approximators \citep{hornik1990universal,hornik1991approximation,Cybenko}. Differently to many traditional methods, neural networks do not require the use of analytic expressions and are not limited to low-dimensional spaces.  On the other hand, training neural networks usually requires very large datasets. While existing datasets for computer vision tasks are numerous and diverse (e.g.~MNIST\footnote{\url{http://yann.lecun.com/exdb/mnist/}}, CIFAR10\footnote{\url{https://www.cs.toronto.edu/~kriz/cifar.html}}, and ImageNet\footnote{\url{https://image-net.org}}), the situation is very different for cosmology. 

Large-volume cosmological hydrodynamic simulations have become a primary tool to study the formation of galaxies and large scale structure \citep{SomervilleDave2015}.  These simulations follow explicitly the evolution of the dark matter and baryonic components in an expanding universe, incorporating a variety of subgrid prescriptions to model key physical processes such as star formation and feedback from massive stars and the impact of Active Galactic Nuclei (AGN) feedback powered by accretion onto supermassive black holes \citep[e.g.][]{illustris:Genel-2014,Angles-Alcazar2017_BHfeedback,Pillepich_2018,SIMBA}.  
However, many of these processes remain poorly understood and require careful tuning of free parameters. While a single simulation can provide a plausible representation of the universe, model-dependent parameter degeneracies limit the predictive power of simulations and their use as primary tool to produce robust constraints on cosmological parameters. 

The Cosmology and Astrophysics with MachinE-learning Simulations (CAMELS) project \citep{CAMELS} has pioneered a new approach, producing thousands of cosmological hydrodynamic simulations to train machine learning algorithms, varying cosmology, initial random field, subgrid galaxy formation model, and stellar and AGN feedback parameters. 

In this paper we present and make publicly available the CAMELS Multifield Dataset, CMD, a large dataset containing hundreds of thousands of 2D maps and 3D grids with different properties obtained from 2,000 distinct universes. Each 2D map and 3D grid represents a region with an area of $(25~h^{-1}{\rm Mpc})^2$ and volume of $(25~h^{-1}{\rm Mpc})^3$, respectively, where many different fields are included, from dark matter to gas and stellar properties, at different redshifts. Each 2D map and 3D grid is associated to a vector of either 2 or 6 numbers: two cosmological parameters (all data) and four astrophysical parameters (only data from hydrodynamic simulations) that define and control the behavior of the parent cosmological simulation in each case. 
The full dataset comprises more than 70 Terabytes and represents the largest collection of 2D maps and 3D grids from state-of-the-art hydrodynamic simulations publicly available to date, and may serve as a standard cosmological and astrophysical dataset to train machine learning models to perform a large variety of tasks.

In our companion papers \citep{baryons_marginalization, Robust_marginalization} we used CMD to perform, for the first time, likelihood-free inference of cosmological parameters at the field level from 2D maps generated from state-of-the-art hydrodynamic simulations of 13 different fields, obtaining very promising results. In this work we describe in detail the architecture and training procedure used in those works. Furthermore, we formulate the problems we encountered as a challenge to the community. We also release all codes and network weights from our companion papers \citep{baryons_marginalization, Robust_marginalization} as a benchmark to compare against.

Information on how to download and manipulate CMD, together with the codes and network weights used for  parameter inference and other tasks can be found at \url{https://camels-multifield-dataset.readthedocs.io}.

This paper is organized as follows. In section \ref{sec:data} we describe CMD in detail. We then outline in section \ref{sec:applications} a few of its possible applications, focusing our attention on parameter inference. We conclude in section \ref{sec:summary}.

\section{Data} 
\label{sec:data}

In this section we describe CMD. We first present CAMELS, the simulation suite used to  generate CMD. We then outline the different fields present in CMD. Next, we explain how the 2D maps and 3D grids were created. Finally, we discuss the labels, symmetries, and storage needs of CMD.

\subsection{The CAMEL Simulations}
\label{sec:simulations}

CMD was generated from CAMELS data \citep{CAMELS}. We now briefly describe those simulations here and refer the reader to \cite{CAMELS} for further details.

CAMELS is a suite of 4,233 numerical simulations: 2,184 state-of-the-art (magneto-)hydrodynamic simulations and 2,049 gravity-only N-body simulations. All simulations follow the evolution of $256^3$ dark matter particles plus $256^3$ fluid elements (only in the case of hydrodynamic simulations) from $z=127$ to the present time, $z=0$, in a periodic volume of $(25~h^{-1}{\rm Mpc})^3$. The initial conditions of all simulations were generated at redshift $z=127$ using second order Lagrangian perturbation theory (2LPT\footnote{\url{https://cosmo.nyu.edu/roman/2LPT/}}). For simplicity, we assumed that the transfer functions of the dark matter and gas fluids of the (magneto-)hydrodynamic simulations were the same and equal to the one of total matter. For each simulation, we saved snapshots at 34 different redshifts, from $z=6$ to $z=0$. To generate CMD, we used the snapshots at $z = [0, 0.5, 1, 1.5, 2]$.

While all simulations follow the evolution of dark matter particles, only the (magneto-)hydrodynamic simulations solve the hydrodynamic equations and implement astrophysical effects such as star formation and feedback from stars and AGN. Each CAMEL simulation belongs to one of three suites: 1) IllustrisTNG (magneto-hydrodynamic simulations), 2) SIMBA (hydrodynamic simulations), and 3) N-body (gravity-only simulations). CMD preserves this naming system. For instance, when addressing the IllustrisTNG neutral hydrogen maps, we refer to the 2D maps containing the neutral hydrogen field that were generated from the CAMELS IllustrisTNG simulation suite. We note that the IllustrisTNG and SIMBA simulations solve the hydrodynamic equations using two completely different methods and implement very different subgrid models.

All simulations share the values of the following cosmological parameters: $\Omega_{\rm b}=0.049$, $h=0.6711$, $n_s=0.9624$, $w=-1$, $\sum m_\nu=0$, $\Omega_K=0$. The values of $\Omega_{\rm m}$ and $\sigma_8$ are however varied across simulations.

The hydrodynamic simulations (IllustrisTNG and SIMBA) also vary four astrophysical parameters, coined $A_{\rm SN1}$, $A_{\rm SN2}$, $A_{\rm AGN1}$, and $A_{\rm AGN2}$ that control the efficiency of supernova and AGN feedback. It is important to emphasize that although the names of these parameters are the same in IllustrisTNG and SIMBA, their implementation and meaning are not identical. This should be taken into account while working with the feedback parameter labels (we provide more details on this in section \ref{subsec:labels}).

Each CAMEL simulation is thus characterized by the suite it belongs to (IllustrisTNG, SIMBA, or N-body) and either 2 or 6 numbers: two cosmological parameters (all simulations) and four astrophysical parameters (only IllustrisTNG and SIMBA). While the astrophysical parameters govern broadly similar quantities in IllustrisTNG and SIMBA, they are implemented differently and hence their absolute values cannot be straightforwardly compared across models. Also, one cannot expect a one-to-one correspondence between the results of a simulation and the values of its parameters, as the simulations are also affected by cosmic variance: the finite volume that the simulations represent does not correspond to a representative sample of the whole universe. Thus, the link between quantities measured from a simulation and the values of its parameters can be considered as probabilistic. 

The CAMELS simulations span a wide range in the values of the cosmological and astrophysical parameters. They were designed precisely in that way to avoid being affected by priors when training neural networks and also to allow overlap in parameter space when using different hydrodynamic simulations. On the one hand, we are mostly interested in performing inference on the value of the cosmological parameters, thus many different models need to be simulated. On the other hand, we know that astrophysical effects can have an impact on cosmological observables but we do not fully understand the physics of those effects, so the most conservative solution is to marginalize over them. For that reason, the values of the astrophysical parameters have to be varied over a wide range to perform a robust marginalization.

Each simulation suite contains four different sets:

\begin{itemize}
\item LH (Latin-Hypercube) is a set of 1,000 simulations for each of the IllustrisTNG and SIMBA suites, where the values of the cosmological and astrophysical parameters are sampled from a latin-hypercube. Each of these simulations has a different initial random seed. The vast majority of 2D maps and 3D grids from CMD were generated from this simulation set.
This is the main simulation set from which we build CMD.
\item 1P (1 parameter at a time) is a set of 61 simulations with the same initial random seed but where only the value of one parameter is varied at a time. CMD contains data from this set.
\item CV (Cosmic Variance) is a set of 27 simulations with the same value of the cosmological and astrophysical parameters but different initial random seeds. A small fraction of 2D maps and 3D grids from CMD were generated from this set. The main purpose of the data from this set is to test the models trained on data from the LH set.
\item EX (Extreme) is a set of 4 simulations covering extreme models. CMD does not contain data from this set.
\end{itemize}

We now describe in more detail the particulars of each simulation type.

\subsubsection{IllustrisTNG}

The IllustrisTNG simulations have been run with the AREPO code\footnote{\url{https://arepo-code.org/}} \citep{Arepo} and employ the same subgrid physics as the original IllustrisTNG simulations \citep{WeinbergerR_16a, PillepichA_16a}. In these simulations, $A_{\rm SN1}$ and $A_{\rm SN2}$ control two properties of galactic winds: the energy emitted per unit of star-formation rate and the wind speed, respectively. $A_{\rm AGN1}$ and $A_{\rm AGN2}$ represent the energy released per unit of black hole accretion rate and the ejection speed (burstiness) for the kinetic mode of black hole feedback. We refer the reader to \cite{CAMELS} for further details on these simulations and their astrophysical parameters.

\subsubsection{SIMBA}

The SIMBA simulations have been run with the \textsc{GIZMO} code\footnote{\url{http://www.tapir.caltech.edu/~phopkins/Site/GIZMO.html}} \citep{Hopkins:2014qka} and employ the same subgrid physics as the original SIMBA simulation \citep{SIMBA}. The parameters $A_{\rm SN1}$ and $A_{\rm SN2}$ control the mass loading factor and speed of galactic winds relative to scalings derived from the FIRE simulations \citep{Muratov2015,Angles-Alcazar2017_BaryonCycle}. The parameter $A_{\rm AGN1}$ determines the momentum flux of kinetic outflows in quasar and jet-mode AGN feedback relative to the black hole accretion rate \citep{Angles-Alcazar2017_BHfeedback} while $A_{\rm AGN2}$ parametrizes the speed of the jet-mode black hole feedback. We refer the reader to \cite{CAMELS} for further details on these simulations and their astrophysical parameters.

We note that although both the IllustrisTNG and SIMBA simulations aim at modeling the properties of cosmic gas, dark matter, and galaxies in a given cosmological model, the way they solve the hydrodynamic equations and implement their subgrid physics is substantially different. Thus, since neither simulation is a priori more accurate than another, this should also be seen as another factor to marginalize over when performing cosmological tasks such as parameter inference. For instance, for inference on the value of the cosmological parameters, it would be desirable that results do not depend on the simulation suite used for training a given model (see \ref{subsec:challenges} for more details). 

\subsubsection{N-body}

The gravity-only simulations have been run with the \textsc{Gadget-III} code \citep{gadget-3_2005}. They only follow the evolution of dark matter particles which represent the cold dark matter plus baryon fluid, but as opposed to the above hydrodynamic simulations, they do not solve the hydrodynamic equations or model astrophysical effects such as supernova and AGN feedback. Thus, the data from these simulations is not affected by astrophysical effects, and therefore, CMD data created from these simulations can be seen as a pristine sample. These simulations only follow the evolution of the total matter field. 

There is one N-body simulation for each hydrodynamical simulation in the IllustrisTNG and SIMBA suites. This is the reason why the number of 2D maps and 3D grids for the total matter field from these simulations is twice as large as those from the IllustrisTNG and SIMBA simulations. In other words, for each total matter map or grid, there is an N-body counterpart.

The N-body simulations use the same random amplitudes and phases for the initial perturbations as their hydrodynamical analogs, such that they follow `the same' universe, except for including only dark matter and not normal matter and correspondingly following only gravity and not other physics.

\begin{table*}[ht]
\begin{center}
\fontsize{7.8pt}{10pt}\selectfont
\renewcommand{\arraystretch}{0.9}
\begin{tabular}{|c|c|c|c|c|c|c|c|c|} 
\hline
\multicolumn{3}{|c|}{ } & \multicolumn{6}{c|}{\textbf{Simulation}} \\ 
\cline{4-9}
\multicolumn{3}{|c|}{ } & \multicolumn{2}{c|}{\textbf{IllustrisTNG}} & \multicolumn{2}{c|}{\textbf{SIMBA}} & \multicolumn{2}{c|}{\textbf{N-body}} \\
\hline 
\textbf{Field} & \textbf{prefix} & \textbf{units} & \textbf{2D maps} & \textbf{3D grids} & \textbf{2D maps} & \textbf{3D grids} & \textbf{2D maps} & \textbf{3D grids} \\
\hline
\hline
Gas density & Mgas & $(h^{-1}M_\odot)/(h^{-1}{\rm kpc})^A$ & 15,000 & 15,000 & 15,000 & 15,000 & - & - \\
\hline
Gas velocity & Vgas & ${\rm km/s}$ & 15,000 & 15,000 & 15,000 & 15,000 & - & - \\
\hline
Gas temperature & T & Kelvin & 15,000 & 15,000 & 15,000 & 15,000 & - & - \\
\hline
Gas pressure & P & $({\rm km/s})(M_\odot/{\rm kpc}^3)$ & 15,000 & 15,000 & 15,000 & 15,000 & - & - \\
\hline
Gas metallicity & Z & - & 15,000 & 15,000 & 15,000 & 15,000 & - & - \\
\hline
Neutral hydrogen density & HI & $h^{-1}M_\odot/(h^{-1}{\rm kpc})^A$ & 15,000 & 15,000 & 15,000 & 15,000 & - & - \\
\hline
Electron number density & ne & $h^{-1}/(h^{-1}{\rm kpc})^A$ & 15,000 & 15,000 & 15,000 & 15,000 & - & - \\
\hline
Magnetic fields & B & Gauss & 15,000 & 15,000 & - & - & - & - \\
\hline
Magnesium over Iron & MgFe & - & 15,000 & 15,000 & 15,000 & 15,000 & - & - \\
\hline
Dark matter density & Mcdm & $h^{-1}M_\odot/(h^{-1}{\rm kpc})^A$ & 15,000 & 15,000 & 15,000 & 15,000 & - & - \\
\hline
Dark matter velocity & Vcdm & km/s & 15,000 & 15,000 & 15,000 & 15,000 & - & - \\
\hline
Stellar mass density & Mstar & $h^{-1}M_\odot/(h^{-1}{\rm kpc})^A$ & 15,000 & 15,000 & 15,000 & 15,000 & - & - \\
\hline
Total matter density & Mtot & $h^{-1}M_\odot/(h^{-1}{\rm kpc})^A$ & 15,000 & 15,000 & 15,000 & 15,000 & $30,000^*$ & $30,000^*$ \\
\hline
\hline
\textbf{Total} &  & & 195,000 & 195,000 & 180,000 & 180,000 & 30,000 & 30,000 \\
\hline
\end{tabular}
\end{center}
\caption{This table shows the number of 2D maps and 3D grids available for the different fields and simulations. All 2D maps are at $z=0$ and contain $256^2$ pixels. The 3D grids contain $128^3$, $256^3$, or $512^3$ voxels and are at redshifts $z=0$, $z=0.5$, $z=1$, $z=1.5$ or $z=2$. We also quote the prefix used for the different fields together with the units used to store the information in the 2D maps and 3D grids. The exponent $A$ that appears in the units of several densities has a value of 2 in the case of 2D maps and of 3 for 3D grids. The symbol $^*$ denotes that for each 2D map and 3D grid of the total matter field of the IllustrisTNG and SIMBA simulations, there is a N-body counterpart.}
\label{tab:data}
\end{table*}

\subsection{Fields}
\label{subsec:fields}

Before describing how we generate the CMD data from the CAMELS simulations, we first outline the different fields that comprise CMD. 

The hydrodynamic simulations contain four different types of particles: 1) gas, 2) dark matter, 3) stars, and 4) black holes. The N-body simulations only have dark matter particles. All particle types have positions, velocities, and masses. The gas particles also contain a set of properties describing the physical state of the gas element they represent: pressure, temperature, metallicity, neutral hydrogen, electron number density, and magnetic fields (only in the case of IllustrisTNG simulations).

Each particle type represents a resolution element of its corresponding component, e.g.~a gas particle represents a fluid element of cosmic gas. The actual 3D shape of these elements can be quite complicated\footnote{For instance, in the case of the IllustrisTNG simulations, the gas particles are the mesh-generating points of a Voronoi tessellation of space.} but we approximate them as uniform spheres for simplicity, characterized by their window function
\begin{equation}
W_{3D}(\vec{x}) = \left\{ 
  \begin{array}{ c l }
    \frac{3}{4\pi R^3} & \quad \textrm{if } |\vec{x}| \leq R \\
    0                 & \quad \textrm{otherwise}
  \end{array}
\right.
\label{eq:window_3D}
\end{equation}
such that 
\begin{equation}
\int_{\vec{x}} W_{3D}(\vec{x}) d^3\vec{x}=1    
\end{equation}
and where $R$ is the radius of the sphere and the integral runs over all 3D space. In the case of gas and dark matter particles, we set this radius to the distance to the 32nd closest\footnote{This number was chosen empirically as a compromise. Higher values will erase small scale features and therefore produce a smoother field than the actual one. On the other hand, smaller values may give rise to sharp transitions and the appearance of noisy features.} gas and dark matter particles, respectively. For stars and black hole particles we set $R=0$, as the resolution of our 2D maps and 3D grids is too coarse to be able to resolve the internal structure of galaxies, i.e.~the window function is just a Dirac delta.

We now describe in more detail each CMD field that represents the spatial distribution of a different property of gas, dark matter, and stars in a given universe. In the following equations for each field, the window function we refer to is the one in Eq. \ref{eq:window_3D} for 3D, and its projection
\begin{equation}
W_{2D}(\vec{x}')=\int_{z} W_{3D}(\vec{x})dz
\end{equation}
when working in 2D (see \ref{subsec:2D_maps} for further details).Table \ref{tab:data} summarizes the fields considered for each simulation suite.

\begin{itemize}
\item \textbf{Gas density.} This field represents the spatial distribution of the density of cosmic gas. To construct 2D maps and 3D grids for this field we need to read the positions and masses of all gas particles in a given simulation. For this field, the quantity stored in every pixel/voxel is the density of gas from all particles that contribute to that location, or more formally
\begin{equation}
    \frac{1}{Q}\sum_i M_{{\rm g},i} \int_{\vec{x}} W_{{\rm g},i}(\vec{x}-\vec{r}_{{\rm g},i}) d\vec{x}~,
\end{equation}
where $W_{{\rm g},i}$, $M_{{\rm g},i}$, and $\vec{r}_{{\rm g},i}$ are the window function, mass, and position of the gas particle $i$. The sum runs over all gas particles and the integral is over the area/volume of a given pixel/voxel. $Q$ represents the area of a pixel in the case of 2D maps or the volume of a voxel for the 3D grids. This field is only present in the IllustrisTNG and SIMBA simulations. 

\item \textbf{Gas velocity.} This field represents the spatial distribution of the modulus of the peculiar velocity vector of cosmic gas $v_{\rm g}=|\vec{v}_{\rm g}|$. To generate 2D maps and 3D grids for this field we need to read the positions, masses, and velocities of all gas particles in a given simulation. The quantity stored in a pixel/voxel is the mass-weighted modulus of the gas velocity from all gas particles contributing to that location, or more formally
\begin{equation}
    \frac{\sum_i M_{{\rm g},i}v_{{\rm g},i}\int_{\vec{x}} W_{{\rm g},i}(\vec{x}-\vec{r}_{{\rm g},i}) d\vec{x}}{\sum_i M_{{\rm g},i} \int_{\vec{x}} W_{{\rm g},i}(\vec{x}-\vec{r}_{{\rm g},i}) d\vec{x}}~,
\end{equation}
where $W_{{\rm g},i}$, $M_{{\rm g},i}$, $v_{{\rm g},i}$, and $\vec{r}_{{\rm g},i}$ are the window function, mass, velocity, and position of the gas particle $i$. The sum runs over all gas particles and the integral is over the area/volume of the considered pixel/voxel. This field is only present in the IllustrisTNG and SIMBA simulations.

We note that the above quantity should not be seen as the gas bulk velocity, which should be computed as
\begin{equation}
    \frac{|\sum_i M_{{\rm g},i}\vec{v}_{{\rm g},i}\int_{\vec{x}} W_{{\rm g},i}(\vec{x}-\vec{r}_{{\rm g},i}) d\vec{x}|}{\sum_i M_{{\rm g},i} \int_{\vec{x}} W_{{\rm g},i}(\vec{x}-\vec{r}_{{\rm g},i}) d\vec{x}}~.
\end{equation}
Instead, in our definition each gas particle makes a positive contribution to the quantity, and its magnitude will be larger for higher velocities, independently of their direction. 

\item \textbf{Gas temperature.} This field represents the spatial distribution of the temperature of the cosmic gas. To generate 2D maps and 3D grids we need to read the positions, masses, and temperatures of all gas particles in a given simulation. The quantity stored in a pixel/voxel is the mass-weighted temperature of cosmic gas from all gas particles contributing to that location, or more formally
\begin{equation}
    \frac{\sum_i M_{{\rm g},i} T_i \int_{\vec{x}} W_{{\rm g},i}(\vec{x}-\vec{r}_{{\rm g},i}) d\vec{x}}{\sum_i M_{{\rm g},i} \int_{\vec{x}} W_{{\rm g},i}(\vec{x}-\vec{r}_{{\rm g},i}) d\vec{x}}
\end{equation}
where $W_{{\rm g},i}$, $M_{{\rm g},i}$, $T_i$, and $\vec{r}_{{\rm g},i}$ are the window function, mass, temperature, and position of the gas particle $i$. The sum runs over all gas particles and the integral is over the area/volume of the considered pixel/voxel. This field is only present in the IllustrisTNG and SIMBA simulations.  

\item \textbf{Gas pressure.} This field represents the spatial distribution of the pressure of the cosmic gas. To generate 2D maps and 3D grids we need to read the positions, masses, and pressures of all gas particles in a given simulation. The quantity stored in a pixel/voxel is the mass-weighted pressure of cosmic gas from all gas particles contributing to that location, or more formally
\begin{equation}
    \frac{\sum_i M_{{\rm g},i}P_i\int_{\vec{x}} W_{{\rm g},i}(\vec{x}-\vec{r}_{{\rm g},i}) d\vec{x}}{\sum_i M_{{\rm g},i} \int_{\vec{x}} W_{{\rm g},i}(\vec{x}-\vec{r}_{{\rm g},i}) d\vec{x}}
\end{equation}
where $W_{{\rm g},i}$, $M_{{\rm g},i}$, $P_i$, and $\vec{r}_{{\rm g},i}$ are the window function, mass, pressure, and position of the gas particle $i$. The sum runs over all gas particles and the integral is over the area/volume of the considered pixel/voxel. This field is only present in the IllustrisTNG and SIMBA simulations.

\item \textbf{Gas metallicity.} This field represents the spatial distribution of the metallicity of the cosmic gas. The metallicity is defined as the ratio between the mass in metals\footnote{In astronomy, metals are all elements heavier than hydrogen and helium.} and the total gas mass: $Z = M_{\rm metal}/M_{\rm g}$. To generate 2D maps and 3D grids we need to read the positions, masses, and metallicities of all gas particles in a given simulation. The quantity stored in a pixel/voxel is the mean metallicity of cosmic gas from all gas particles contributing to that location, or more formally
\begin{equation}
    \frac{\sum_i M_{{\rm g},i} Z_i \int_{\vec{x}} W_{{\rm g},i}(\vec{x}-\vec{r}_{{\rm g},i}) d\vec{x}}{\sum_i M_{{\rm g},i} \int_{\vec{x}} W_{{\rm g},i}(\vec{x}-\vec{r}_{{\rm g},i}) d\vec{x}}
\end{equation}
where $W_{{\rm g},i}$, $M_{{\rm g},i}$, $Z_i$, and $\vec{r}_{{\rm g},i}$ are the window function, mass, metallicity, and position of the gas particle $i$. The sum runs over all gas particles and the integral is over the area/volume of the considered pixel/voxel. This field is only present in the IllustrisTNG and SIMBA simulations.

\item \textbf{Neutral hydrogen density.} This field represents the spatial distribution of the density of neutral hydrogen. To generate 2D maps and 3D grids we need to read the positions and neutral hydrogen masses of all gas particles in a given simulation. The quantity stored in a pixel/voxel is the total neutral hydrogen density from all gas particles contributing to that location, or more formally
\begin{equation}
    \frac{1}{Q}\sum_i M_{{\rm HI},i} \int_{\vec{x}} W_{{\rm g},i}(\vec{x}-\vec{r}_{{\rm g},i}) d\vec{x}~,
\end{equation}
where $W_{{\rm g},i}$, $M_{{\rm HI},i}$, and $\vec{r}_{{\rm g},i}$ are the window function, neutral hydrogen mass, and position of the gas particle $i$. The sum runs over all gas particles and the integral is over the area/volume of the considered pixel/voxel. $Q$ represents the area of a pixel in the case of 2D maps or the volume of a voxel for the 3D grids.  This field is only present in the IllustrisTNG and SIMBA simulations.

\item \textbf{Electron number density.} This field represents the spatial distribution of the density of electrons. To generate 2D maps and 3D grids we need to read the positions and electron abundances of all gas particles in a given simulation. The quantity stored in a pixel/voxel is the total electron number density from all gas particles contributing to that location, or more formally
\begin{equation}
    \frac{1}{Q}\sum_i n_{e,i} \int_{\vec{x}} W_{{\rm g},i}(\vec{x}-\vec{r}_{{\rm g},i}) d\vec{x}~,
\end{equation}
where $W_{{\rm g},i}$, $n_{e,i}$, and $\vec{r}_{{\rm g},i}$ are the window function, number of electrons, and position of the gas particle $i$. The sum runs over all gas particles and the integral is over the area/volume of the considered pixel/voxel. $Q$ represents the area of a pixel in the case of 2D maps or the volume of a voxel for the 3D grids. This field is only present in the IllustrisTNG and SIMBA simulations.

\item \textbf{Magnetic fields.} This field represents the spatial distribution of the magnitude of the magnetic field vector of the cosmic gas: $B=|\vec{B}|$. To generate 2D maps and 3D grids we need to read the positions, masses, and magnetic field modulus of all gas particles in a given simulation. The quantity stored in a pixel/voxel is the mass-weighted modulus of the magnetic field vector from all gas particles contributing to that location, or more formally
\begin{equation}
    \frac{\sum_i M_{{\rm g},i} B_i \int_{\vec{x}} W_{{\rm g},i}(\vec{x}-\vec{r}_{{\rm g},i}) d\vec{x}}{\sum_i M_{{\rm g},i} \int_{\vec{x}} W_{{\rm g},i}(\vec{x}-\vec{r}_{{\rm g},i}) d\vec{x}}
\end{equation}
where $W_{{\rm g},i}$, $B_i$, and $\vec{r}_{{\rm g}, i}$ are the window function, magnetic field modulus, and position of the gas particle $i$. The sum runs over all gas particles and the integral is over the area/volume of the considered pixel/voxel. This field is only present in the IllustrisTNG simulations.

\item \textbf{Magnesium over Iron ratio.} This field represents the spatial distribution of the ratio between the magnesium and iron masses of the cosmic gas. The mass of each gas particle represents the sum of the mass in hydrogen, helium, and metals of that particle. Among the metals, our simulations track the masses of the carbon, nitrogen, oxygen, magnesium, silicon, and iron elements. This field thus represents the ratio between the masses of those two elements that belong to each gas particle. To generate 2D maps and 3D grids we need to read the positions, magnesium masses, and iron masses of all gas particles in a given simulation. The quantity stored in a pixel/voxel is the ratio between all magnesium and iron masses from all gas particles contributing to that location, or more formally
\begin{equation}
    \frac{\sum_i M_{{\rm Mg},i} \int_{\vec{x}} W_{{\rm g},i}(\vec{x}-\vec{r}_{{\rm g},i}) d\vec{x}}{\sum_i M_{{\rm Fe},i} \int_{\vec{x}} W_{{\rm g},i}(\vec{x}-\vec{r}_{{\rm g},i}) d\vec{x}}
\end{equation}
where $W_{{\rm g},i}$, $M_{{\rm Mg},i}$, $M_{{\rm Fe},i}$, and $\vec{r}_{{\rm g},i}$ are the window function, magnesium mass, iron mass, and position of the gas particle $i$. The sum runs over all gas particles and the integral is over the area/volume of the considered pixel/voxel. This field is only present in the IllustrisTNG and SIMBA simulations.

\item \textbf{Dark matter density.} This field represents the spatial distribution of the dark matter density. To generate 2D maps and 3D grids we need to read the positions and masses of all dark matter particles in a given simulation. The quantity stored in a pixel/voxel is the dark matter density from all dark matter particles contributing to that location, or more formally
\begin{equation}
    \frac{1}{Q}\sum_i M_{{\rm dm},i} \int_{\vec{x}} W_{{\rm dm},i}(\vec{x}-\vec{r}_{{\rm dm},i}) d\vec{x}~,
\end{equation}
where $W_{{\rm dm},i}$, $M_{{\rm dm},i}$, and $\vec{r}_{{\rm dm},i}$ are the window function, mass, and position of the dark matter particle $i$. The sum runs over all dark matter particles and the integral is over the area/volume of the considered pixel/voxel. $Q$ represents the area of a pixel in the case of 2D maps or the volume of a voxel for the 3D grids. This field is only present in the IllustrisTNG and SIMBA simulations\footnote{Note, however, that for the N-body simulations the definitions of the dark matter mass  field and the total matter field (see below), which does exist for those simulations, are by construction identical.}.

\item \textbf{Dark matter velocity.} This field represents the spatial distribution of the modulus of the peculiar velocity vector of the dark matter $v_{\rm dm}=|\vec{v}_{\rm dm}|$. To generate 2D maps and 3D grids for this field we need to read the positions, masses, and velocities of all dark matter particles in a given simulation. The quantity stored in a pixel/voxel is the mass-weighted modulus of the dark matter velocity from all dark matter particles contributing to that location, or more formally
\begin{equation}
    \frac{\sum_i M_{{\rm dm},i}v_{{\rm dm},i}\int_{\vec{x}} W_{{\rm dm},i}(\vec{x}-\vec{r}_{{\rm dm},i}) d\vec{x}}{\sum_i M_{{\rm dm},i} \int_{\vec{x}} W_{{\rm dm},i}(\vec{x}-\vec{r}_{{\rm dm},i}) d\vec{x}}~,
\end{equation}
where $W_{{\rm dm},i}$, $M_{{\rm dm},i}$, $v_{{\rm dm},i}$, and $\vec{r}_{{\rm dm},i}$ are the window function, mass, velocity, and position of the dark matter particle $i$. The sum runs over all dark matter particles and the integral is over the area/volume of the considered pixel/voxel. This field is only present in the IllustrisTNG and SIMBA simulations.

We note that the above quantity should not be seen as the dark matter bulk velocity, which should be computed as
\begin{equation}
    \frac{|\sum_i M_{{\rm dm},i}\vec{v}_{{\rm dm},i}\int_{\vec{x}} W_{{\rm dm},i}(\vec{x}-\vec{r}_{{\rm dm},i}) d\vec{x}|}{\sum_i M_{{\rm dm},i} \int_{\vec{x}} W_{{\rm dm},i}(\vec{x}-\vec{r}_{{\rm dm},i}) d\vec{x}}~.
\end{equation}
Instead, in our definition each gas particle has a positive contribution and its magnitude will be larger for higher velocities, independently of their direction.

\item \textbf{Stellar mass density.} This field represents the spatial distribution of the stellar mass density. To generate 2D maps and 3D grids we need to read the positions and masses of all star particles in a given simulation. The quantity stored in a pixel/voxel is the stellar mass density from all star particles contributing to that location, or more formally
\begin{equation}
    \frac{1}{Q}\sum_i M_{*,i} \int_{\vec{x}} W_{*,i}(\vec{x}-\vec{r}_{*,i}) d\vec{x}~,
\end{equation}
where $W_{*,i}$, $M_{*,i}$, and $\vec{r}_{*,i}$ are the window function, mass, and position of the star particle $i$. The sum runs over all star particles and the integral is over the area/volume of the considered pixel/voxel. $Q$ represents the area of a pixel in the case of 2D maps or the volume of a voxel for the 3D grids. This field is only present in the IllustrisTNG and SIMBA simulations.

\item \textbf{Total matter mass.} This field represents the spatial distribution of the total matter mass. Total matter is defined as the sum of gas, dark matter, stars, and black holes and thus represents the total amount of mass, baryonic as well as dark, in a given universe. To generate 2D maps and 3D grids we need to read the positions and masses of all gas, dark matter, stars, and black holes particles in a given simulation. The quantity stored in a pixel/voxel is the total matter density from all different particles contributing to that location, or more formally
\begin{eqnarray}
    \frac{1}{Q}[&\sum_i& M_{{\rm g},i} \int_{\vec{x}} W_{{\rm g},i}(\vec{x}-\vec{r}_{{\rm g},i}) d\vec{x} \nonumber \\ 
  + &\sum_i& M_{{\rm dm},i} \int_{\vec{x}} W_{{\rm dm},i}(\vec{x}-\vec{r}_{{\rm dm},i}) d\vec{x} \nonumber \\
  + &\sum_i& M_{*,i} \int_{\vec{x}} W_{*,i}(\vec{x}-\vec{r}_{*,i}) d\vec{x} \nonumber \\
  + &\sum_i& M_{{\rm bh},i} \int_{\vec{x}} W_{{\rm bh},i}(\vec{x}-\vec{r}_{{\rm bh},i}) d\vec{x}]~,
\end{eqnarray}
where $W_{{\rm g},i}$, $W_{{\rm dm},i}$, $W_{*,i}$, and $W_{{\rm bh},i}$ are the window function of $i$ gas, dark matter, star, and black hole particle, respectively. $M_{{\rm g},i}$, $M_{{\rm dm},i}$, $M_{*,i}$, $M_{{\rm bh},i}$ represent the mass of the $i$ gas, dark matter, star, and black hole particle, correspondingly. $\vec{r}_{{\rm g},i}$, $\vec{r}_{{\rm dm},i}$, $\vec{r}_{*,i}$, $\vec{r}_{{\rm bh},i}$ are the position of the $i$ gas, dark matter, star, and black hole particle. The sums run over all particles of the different types and the integral is over the area/volume of the considered pixel/voxel. $Q$ represents the area of a pixel in the case of 2D maps or the volume of a voxel for the 3D grids.

We note that in the case of the N-body simulations, we only need to read the positions and masses of the dark matter particles and evaluate the second term of the above expression. This field is present in all three simulation types: IllustrisTNG, SIMBA, and N-body. We emphasize that the number of 2D maps and 3D grids from the N-body simulations is equal to the sum of those from the IllustrisTNG and SIMBA simulations, since for each map and grid from these simulations there is an N-body counterpart.

\end{itemize}

\subsection{2D Maps}
\label{subsec:2D_maps}

We now describe the method we use to generate the 2D maps and their characteristics. In Table \ref{tab:data} we summarize the number and properties of the CMD 2D maps.

The 2D maps are generated as follows. First, we consider a given simulation and read the positions and properties of the considered field (see the above subsection for further details on each field). Next, we compute the radius of the considered particles as the distance to the 32nd closest gas and dark matter particle (in the case of gas and dark matter particles, respectively), or set it to zero for star and black hole particles. We then consider a slice of dimensions $25\times25\times5~(h^{-1}{\rm Mpc})^3$ and select the particles that lie inside it. For each simulation we take 15 slices: 5 in the XY plane, 5 in the XZ plane, and 5 in the YZ plane. We note that slices with the same projection direction do not overlap in space. We then project the window function of the considered particles into 2D:
\begin{equation}
W_{2D}(r,\theta)=\int_z W_{3D}(r,\theta,z) dz    
\end{equation}
where we have made use of cylindrical coordinates, and $z$ represents the axes along which we project the slice. Note that by construction, the quantities will be preserved in the projection, i.e.
\begin{eqnarray}
    &&\int_{\vec{x}'} W_{2D}(r,\theta)d\vec{x}'=\int_r \int_\theta W_{2D}(r,\theta) dr d\theta = \\ 
    &&\int_r \int_\theta \int_z W_{3D}(r,\theta,z)drd\theta dz=\int_{\vec{x}} W_{3D}(r,\theta,z)d\vec{x}~. \nonumber
\end{eqnarray}
Finally, we deposit the properties associated to these circles into a regular 2D grid with $256\times256$ pixels by performing the integrals of section \ref{subsec:fields}. We do this numerically, by sampling each circle with 1,000 tracers that are distributed such that all of them cover the same area, and assign the contribution of each tracer to the pixel that contains its center. Although this procedure is approximate, in the limit that the number of tracers tends to infinity, one recovers the correct exact result of depositing circles into a 2D regular grid. We have checked that with 1,000 tracers our results are converged. We note that more accurate results can be obtained, if needed, from the 3D grids, whose constructions follows a very different procedure to the 2D maps (see section \ref{subsec:3D_grids} for more details).

All 2D maps are created at $z=0$ and contain $256\times256$ pixels over an area of $25\times25~(h^{-1}{\rm Mpc})^2$. For each field we generate 15,000 maps: 15 maps per simulation for 1,000 simulations. Each 2D map is described by either two or six numbers: two cosmological parameters (all maps) and four astrophysical parameters (only the maps generated from the IllustrisTNG and SIMBA simulations).

The reason why CMD only contains 2D maps at $z=0$ is because it is possible to generate maps from the 3D grids (see \ref{subsec:3D_grids}), and also to allow users to fully reproduce the results obtained for the parameter inference task described below. For instance, one can take a slice of a 3D grid and project it into a 2D plane. From the 3D grids we can generate 2D maps with different widths, at different resolutions, and different redshifts. Furthermore, in some cases one may want to use 2D maps that partially overlap in the projected direction. We thus recommend readers to use the 3D grids to generate 2D maps that fulfill their needs. 

\begin{figure*}
\centering
  \includegraphics[width=0.99\linewidth]{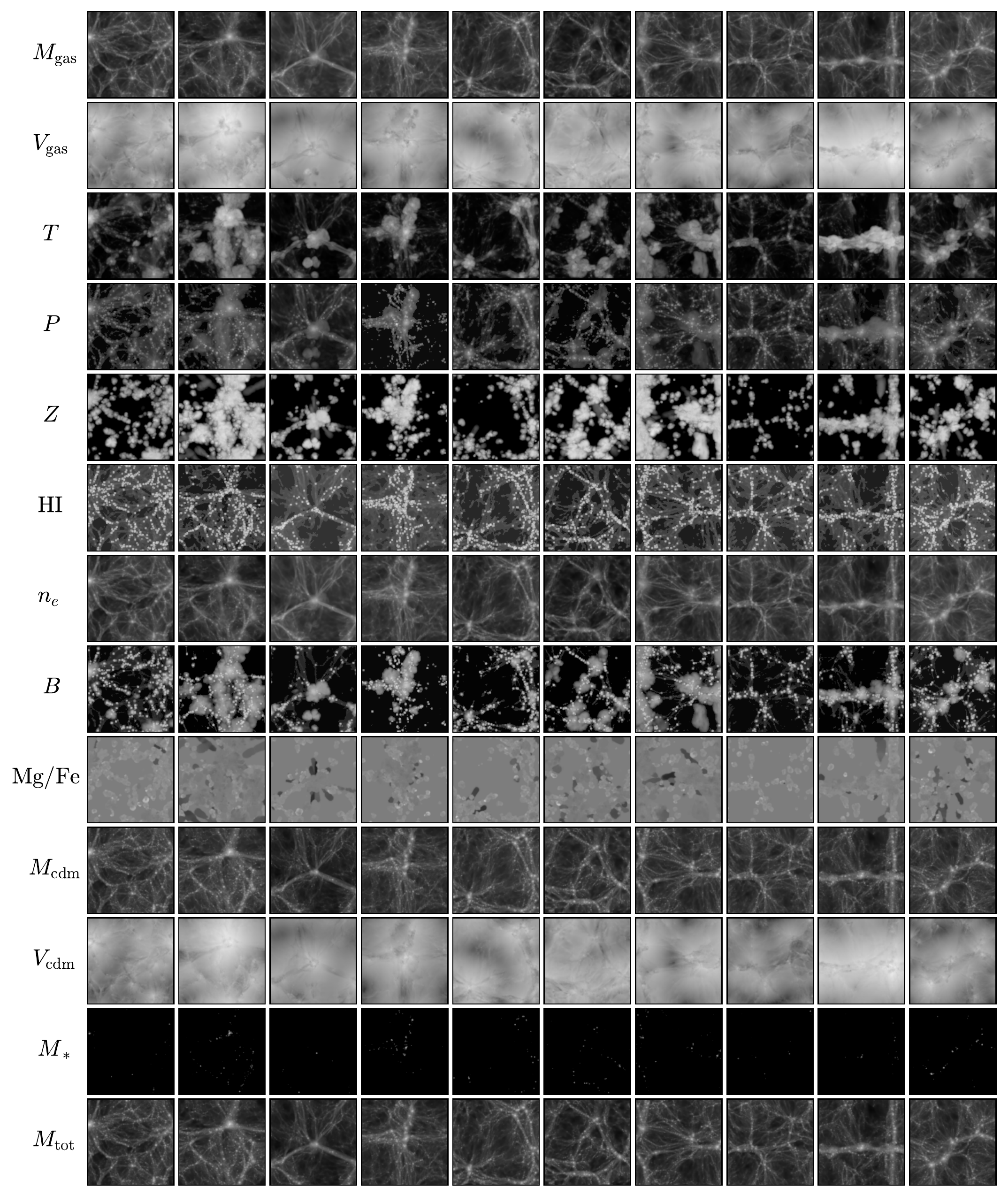}
  \vspace{-0.3cm}
\caption{We show 10 examples of 2D maps for each of the 13 different fields present in the IllustrisTNG simulations. Each image has a different value of the cosmological parameters, $\Omega_{\rm m}$ and $\sigma_8$, and astrophysical parameters $A_{\rm SN1}$, $A_{\rm SN2}$, $A_{\rm AGN1}$, $A_{\rm AGN2}$. The different columns represent the same region for the different fields. Each image has a physical size of $25\times25~(h^{-1}{\rm Mpc})^2$.}
\label{fig:maps}
\end{figure*}

We note that different fields from the same simulation may exhibit a very tight spatial correlation on large scales. This can be seen in Fig. \ref{fig:maps}, where we show an example of CMD maps of the IllustrisTNG simulations for all fields. In the online documentation we show examples of how to read and manipulate the files containing the 2D maps. We also describe there how to generate 2D maps from the 3D grids.

\subsection{3D grids}
\label{subsec:3D_grids}

We now describe the method we employ to generate the 3D grids and present their features. Table \ref{tab:data} shows the characteristics of the CMD 3D grids.

The 3D grids are constructed as follows. First, we read the positions and properties of the considered particles (see \ref{subsec:fields} for details). Next, we compute the radius of the considered particles as the distance to the 32nd closest gas and dark matter particle (in the case of gas and dark matter particles, respectively), or set it to zero for star and black hole particles. Next, we deposit the properties sampled by the spheres into a regular 3D grid with $128^3$, $256^3$, or $512^3$ voxels using \textsc{voxelize}\footnote{\url{https://github.com/leanderthiele/voxelize}}, a code that calculates the integrals of section \ref{subsec:fields} in a precise way, making use of the \textsc{overlap} code \citep{Strobl2016Overlap}\footnote{\url{https://github.com/severinstrobl/overlap}}.
We emphasize that the window function of the particles is given by Eq. \ref{eq:window_3D} and is not the same for different particles.
This means that the window function cannot be trivially removed from power spectra in order to enable direct comparison to theory predictions. However, for machine learning applications this is not an issue, and we believe that the choice of spherical kernels with physically motivated radii leads to smoother fields which may benefit training of convolutional neural networks. 

The 3D grids are constructed at redshifts 0, 0.5, 1, 1.5, and 2. For each simulation, field, and redshift, CMD provides 3 grids with $128^3$, $256^3$, and $512^3$ voxels. We note that these three grids represent exactly the same field at different resolutions. All grids cover a comoving periodic volume of $(25~h^{-1}{\rm Mpc})^3$. Different fields from the same simulation may exhibit a very tight spatial correlation on large scales, as in the case of the 2D maps.

In the online documentation we provide examples illustrating how to read the files containing the 3D fields and how to manipulate the data.

\subsection{Labels}
\label{subsec:labels}

The 2D maps and 3D grids from the IllustrisTNG and SIMBA simulations are characterized by six numbers: two cosmological parameters ($\Omega_{\rm m}$ and $\sigma_8$), and four astrophysical parameters ($A_{\rm SN1}$, $A_{\rm SN2}$, $A_{\rm AGN1}$, $A_{\rm AGN2}$). In the case of maps and grids from the N-body simulations, the labels are only $\Omega_{\rm m}$ and $\sigma_8$. By construction, each parameter can vary within a wide range; many values are so extreme that the corresponding universes simulated in CAMELS are far from reality. This is however not a problem as we want our results to be unaffected by our priors \citep{Villaescusa-Navarro_2020c}. In Table \ref{tab:labels} we show the range of variation of each parameter together with its distribution.

\begin{table}
\begin{center}
\renewcommand{\arraystretch}{0.9}
\begin{tabular}{|c|c|c|} 
\hline 
\textbf{Parameter} & \textbf{Range of variation} & \textbf{Distribution}  \\
\hline \hline
$\Omega_{\rm m}$ & 0.1 - 0.5 & uniform \\
\hline
$\sigma_8$ & 0.6 - 1.0 & uniform\\
\hline
$A_{\rm SN1}$ & 0.25 - 4.0 & log uniform\\
\hline
$A_{\rm SN2}$ & 0.5 - 2.0 & log uniform\\
\hline
$A_{\rm AGN1}$ & 0.25 - 4.0 & log uniform\\
\hline
$A_{\rm AGN2}$ & 0.5 - 2.0 & log uniform\\
\hline
\end{tabular}
\end{center}
\caption{Each 2D map and 3D grid of CMD is characterized by six numbers (two in the case of data from the N-body simulations). This table shows the range of variation and the distribution of each parameter. log uniform means that the distribution of the parameter is uniform when the logarithm of the value is taken.}
\label{tab:labels}
\end{table}

While the values of the cosmological parameters represent the same property in all simulations, 2D maps, and 3D grids, the same is not true for the astrophysical parameters, whose definition and implementation is different in the IllustrisTNG and SIMBA simulations. Thus, if a neural network is trained to infer the value of $\Omega_{\rm m}$ and $\sigma_8$ from 2D maps of  IllustrisTNG simulations, the same network can be tested on 2D maps from the SIMBA and N-body simulations to see if it is able to recover the values of those parameters. On the other hand, a network trained on, e.g., SIMBA 3D grids to infer the values of the astrophysical parameters, is not expected to work when tested in 3D grids from the IllustrisTNG simulations.

For cosmological applications, in general, the astrophysical parameters and their effects should be taken as nuisance parameters to be marginalized over.

\subsection{Symmetries}
\label{subsec:symmetries}

It is important to know the symmetries of the CMD data to exploit them, or to enforce them, when using machine learning models or other methods.

The simulations are equivariant under rotations around integer multiples of $\pi/2$ (equivariance for other rotation angles holds approximately), parity, and translations. In order to implement the latter, periodic boundary conditions apply (not for cropped regions, of course).
This information can be useful in some cases. For instance, instead of padding convolutional layers with zeros, one can use periodic padding for the same task, which may improve the performance of the model.

\subsection{Disk space}
\label{subsec:disk}

We now briefly describe the disk space needed to store the different CMD elements. The property stored in every pixel of a 2D map and in every voxel of a 3D grid is a float that takes 4 bytes. Thus, a 2D map occupies $256\times256\times4=0.25$ MB. In CMD, every field map contains 15,000 maps, so these files will require $3.7$ GB per field. Since CMD has 27 different files for the 2D maps (13 fields for IllustrisTNG, 12 fields for SIMBA, and 1+1 field for N-body\footnote{We do not store the 30,000 maps in a single file, but in two in order to facilitate the correspondence with the maps from the IllustrisTNG and SIMBA maps}), storing all CMD 2D maps will require $\sim100$ GB.

The files hosting the 3D grids instead contain 1,000 grids per field, so each of those field files will occupy $N^3\times4\times1000$ bytes, i.e.~7.8 GB ($N=128$), 62.5 GB ($N=256$), 500 GB ($N=512$). Hence, the 27 field files and three different resolutions combined will require 15 TB for all 3D grids at a single redshift. All CMD grids, at the 5 different redshifts require 75 TB.

We note that in the future we will generate more 2D maps and 3D grids at additional redshifts. The online documentation will always be updated to reflect any new data that is not described in this paper.

\section{Applications}
\label{sec:applications}

In this section we outline a few possible tasks that can be carried out with CMD. The list of applications discussed below is far from comprehensive, but is rather just a subset of all possible applications that can be carried out with such a rich and complex dataset.

\subsection{Parameter inference}
\label{subsec:inference}

One of the main applications of CMD is parameter inference: given a 2D map or a 3D grid, $\textbf{X}$, develop a method that predicts the posterior of the parameters
\begin{equation}
p(\vec{\theta}|{\bf X})
\end{equation}
where $\vec{\theta}$ can be a single parameter, e.g.~$\Omega_{\rm m}$, or several or them such as $\vec{\theta}=\{\Omega_{\rm m}, \sigma_8\, A_{\rm SN1}, A_{\rm SN2}, A_{\rm AGN1}, A_{\rm AGN2}\}$. We emphasize that due to cosmic variance, i.e.~due to the finite volume covered in the simulations, no one-to-one correspondence between the 2D maps or 3D grids and the values of the parameters exists. The inference can be carried out from a single field, or several fields can be used together in what is called a \textit{multifield}.

Being able to extract cosmological information, at the field level, from 2D maps and 3D grids while marginalizing over astrophysical effects is one of the main goals of modern cosmology. CMD represents a state-of-the-art dataset that is optimized for this task. In our companion papers \citep{baryons_marginalization, Robust_marginalization} we have made use of CMD to perform this task for the first time. Here, we now describe in detail the architecture and training procedure used in those works, and describe the problems we encountered as a challenge to the community. We note that in those works we only used the 2D maps, so all the details below apply to this data and not to the 3D grids. We also make publicly available all codes and network weights for this task at \url{https://camels-multifield-dataset.readthedocs.io}. 

The way we carried out the inference in our companion papers \citep{baryons_marginalization, Robust_marginalization} was to predict the mean and standard deviation of the marginal posterior for each parameter. Thus, given a 2D map, the network will output two numbers for each parameter. 

\subsubsection{Architecture}
\label{subsec:architecture}

The architecture of our model consists of a set of convolutional layers (CNNs) followed by a fully connected layer. In detail:

\begin{enumerate}
\item \textbf{Input:} $C\times256\times256 \rightarrow$
\vspace{0.2cm} \hrule
\item  CNN (3, 1, 1) $\rightarrow 2H\times256\times256$ $\rightarrow$
\item LeakyReLU $\rightarrow$
\item CNN(3,1,1) $\rightarrow 2H\times256\times256$ $\rightarrow$
\item BatchNorm $\rightarrow$ LeakyReLU $\rightarrow$
\item CNN(2,2,0) $\rightarrow 2H\times128\times128$ $\rightarrow$
\item BatchNorm $\rightarrow$ LeakyReLU $\rightarrow$
\vspace{0.2cm} \hrule

\item CNN(3,1,1) $\rightarrow 4H\times128\times128$ $\rightarrow$
\item BatchNorm $\rightarrow$ LeakyReLU $\rightarrow$
\item CNN(3,1,1) $\rightarrow 4H\times128\times128$ $\rightarrow$
\item BatchNorm $\rightarrow$ LeakyReLU $\rightarrow$
\item CNN(2,2,0) $\rightarrow 4H\times64\times64$ $\rightarrow$
\item BatchNorm $\rightarrow$ LeakyReLU $\rightarrow$
\vspace{0.2cm} \hrule

\item CNN(3,1,1) $\rightarrow 8H\times64\times64$ $\rightarrow$
\item BatchNorm $\rightarrow$ LeakyReLU $\rightarrow$
\item CNN(3,1,1) $\rightarrow 8H\times64\times64$ $\rightarrow$
\item BatchNorm $\rightarrow$ LeakyReLU $\rightarrow$
\item CNN(2,2,0) $\rightarrow 8H\times32\times32$ $\rightarrow$
\item BatchNorm $\rightarrow$ LeakyReLU $\rightarrow$
\vspace{0.2cm} \hrule

\item CNN(3,1,1) $\rightarrow 16H\times32\times32$ $\rightarrow$
\item BatchNorm $\rightarrow$ LeakyReLU $\rightarrow$
\item CNN(3,1,1) $\rightarrow 16H\times32\times32$ $\rightarrow$
\item BatchNorm $\rightarrow$ LeakyReLU $\rightarrow$
\item CNN(2,2,0) $\rightarrow 16H\times16\times16$ $\rightarrow$
\item BatchNorm $\rightarrow$ LeakyReLU $\rightarrow$
\vspace{0.2cm} \hrule

\item CNN(3,1,1) $\rightarrow 32H\times16\times16$ $\rightarrow$
\item BatchNorm $\rightarrow$ LeakyReLU $\rightarrow$
\item CNN(3,1,1) $\rightarrow 32H\times16\times16$ $\rightarrow$
\item BatchNorm $\rightarrow$ LeakyReLU $\rightarrow$
\item CNN(2,2,0) $\rightarrow 32H\times8\times8$ $\rightarrow$
\item BatchNorm $\rightarrow$ LeakyReLU $\rightarrow$
\vspace{0.2cm} \hrule

\item CNN(3,1,1) $\rightarrow 64H\times8\times8$ $\rightarrow$
\item BatchNorm $\rightarrow$ LeakyReLU $\rightarrow$
\item CNN(3,1,1) $\rightarrow 64H\times8\times8$ $\rightarrow$
\item BatchNorm $\rightarrow$ LeakyReLU $\rightarrow$
\item CNN(2,2,0) $\rightarrow 64H\times4\times4$ $\rightarrow$
\item BatchNorm $\rightarrow$ LeakyReLU $\rightarrow$
\vspace{0.2cm} \hrule

\item CNN(4,1,0) $\rightarrow 128H\times1\times1$ $\rightarrow$
\item BatchNorm $\rightarrow$ LeakyReLU $\rightarrow$
\vspace{0.2cm} \hrule

\item Flatten tensor $\rightarrow$ $128H$ $\rightarrow$
\item Dropout (DR) $\rightarrow$
\item FC(128H, 64H) $\rightarrow 64H$
\item LeakyReLU $\rightarrow$ Dropout (DR) $\rightarrow$
\item FC(64H, 12) $\rightarrow$ 12
\vspace{0.2cm} \hrule
\item \textbf{Output:} mean + std posterior (12 numbers)

\end{enumerate}
where $C$ is the number of channels of the input map; for single fields $C=1$ while this number is larger than one in the case of multifield. $H$ is a hyper-parameter that controls the number of channels in the different CNNs; larger values will increase the number of channels and therefore make the network more complex. DR is the dropout rate, which is another hyper-parameter of the network. The notation CNN(K,S,P) indicates a CNN layer with kernel size K, strides S, and padding P. The input and output number of channels can be inferred from the scheme. E.g.~the fourth CNN (step 8 above) takes as input $2H$ channels of images having $128\times128$ pixels and returns $4H$ channels of images that have $128\times128$ pixels. All CNN layers use periodic padding.

FC(A,B) indicates a fully connected layer with A and B input and output values, respectively. The model has twelve outputs, corresponding to the mean and standard deviation of the marginal posterior for each of the six cosmological and astrophysical parameters. In the case of N-body maps, we set the number of output features to 4, since these maps are fully characterized by $\Omega_{\rm m}$ and $\sigma_8$.

\subsubsection{Loss function}
\label{subsec:loss}

The above model outputs 2 numbers per parameter, the mean ($\mu_i$) and standard deviation ($\sigma_i$) of the marginal posterior:
\begin{eqnarray}
\mu_i(\textbf{X}) &=& \int_{\theta_i} p(\theta_i | \textbf{X}) \theta_i d\theta_i~,\\
&& \nonumber \\
\sigma_i(\textbf{X}) &=& \int_{\theta_i} p(\theta_i | \textbf{X}) (\theta_i - \mu_i)^2 d\theta_i~,
\label{Eq:mean_posterior}
\end{eqnarray}
where $p(\theta_i|\textbf{X})$ is the marginal posterior over the parameter $i$
\begin{equation}
p(\theta_i|\textbf{X}) = \int_{\theta} p(\theta_1,\theta_2,...\theta_n | \textbf{X})d\theta_1...d\theta_{i-1}d\theta_{i+1}...d\theta_n~.
\label{Eq:variance_posterior}
\end{equation}
In this notation, $\textbf{X}$ represents a 2D map. Following the moments network work presented in \cite{Niall_2020}, we define the loss function such that the output of the network converges to the above quantities:
\begin{eqnarray}
\mathcal{L}&=&\sum_{i=1}^6\log\left(\sum_{j\in{\rm batch}}\left(\theta_{i,j} - \mu_{i,j}\right)^2\right)\nonumber \\
+&&\sum_{i=1}^6\log\left(\sum_{j\in{\rm batch}}\left(\left(\theta_{i,j} - \mu_{i,j}\right)^2 - \sigma_{i,j}^2 \right)^2\right)~.
\label{Eq:loss}
\end{eqnarray}

We note that this loss function differs from the original one presented in \cite{Niall_2020},
\begin{equation}
\mathcal{L}=\sum_{i=1}^6\sum_{j\in{\rm batch}}\left(\left(\theta_{i,j} - \mu_{i,j}\right)^2 + \left(\left(\theta_{i,j} - \mu_{i,j}\right)^2 - \sigma_{i,j}^2 \right)^2\right)~.
\label{Eq:loss_Jeffrey}
\end{equation}
We replace the arithmetic sum by the sum of the logarithm of each term (either posterior mean or posterior standard deviation). 
Empirically, we have found that our modified loss function provides much tighter and reliable values for $\mu_i$ and $\sigma_i$ than its original version.

The reason for this is that different parameters can be constrained more accurately than others by the network. For instance, there are fields where the impact of astrophysical processes is very mild, while they are very sensitive to cosmology (e.g.~the dark matter field). In those cases, the loss function in Eq.~\ref{Eq:loss_Jeffrey} will be eventually dominated by the contribution from the astrophysical parameters, preventing the network from further improving constraints on the cosmological parameters. 

By taking the logarithm of each term, before the sum, as in Eq.~\ref{Eq:loss}, we are effectively multiplying the losses of all terms instead of summing them, which prevents the problem mentioned above from occurring. A different way to see this is that when taking the logarithm of the terms, we are rescaling all terms to the same order of magnitude, and therefore the loss function will give a similar weight to all terms. 

\subsubsection{Training procedure}
\label{subsec:training}

Each field contains 15,000 2D maps that we split into training (13,500 maps), validation (750 maps), and testing (750 maps) sets. This split is not done randomly from the maps themselves, but rather based on the simulations they were generated from, such that maps that have the same value of the cosmological and astrophysical parameters all belong to only one of these three sets. Thus, we take maps from 900, 50, and 50 simulations for training, validation, and testing, respectively, which amounts to the numbers of maps above given that there exist 15 maps for each simulation. We do this in this way to avoid hidden correlations between maps from the same simulation that we do not want the network to learn.

Our model has four hyper-parameters: 1) the learning rate (lr), 2) the weight decay (wd), 3) the number of channels in the CNNs (H), and 4) the dropout rate of the fully connected layers (DR). For a given value of the hyper-parameters, we train the above network by minimizing the loss function of Eq. \ref{Eq:loss} using gradient descent. We employ the AdamW optimizer \citep{AdamW} with betas equal to 0.5 and 0.999.\footnote{See,  e.g., \cite{Goodfellow-et-al-2016}, for more information on these and other common deep learning concepts.}

We train using a cyclicLR scheduler \citep{CyclicLR} with a minimum learning rate of $10^{-9}$ and a maximum equal to lr. We take 500 steps up and another 500 steps down to define the period of the scheduler. We use a batch size of 128 and train the model for 200 epochs. We save the weights of the model with the lowest validation loss. 

The value of the hyper-parameters is optimized using the \textsc{optuna} package \citep{Optuna}. For each field, we consider at least 50 trials, where a trial represents the result of training the network with a given value of the hyper-parameters. Thus, for each field, we save the weights of at least 50 different models. 

\textsc{Optuna} produces a database with the information of every trial, such as trial number, value of the hyper-parameters, validation loss, etc. In the online documentation we explain how to read these databases together with the files containing the weights of the networks. With those files in hand, the model can be tested on either the test set or a new dataset.

\subsubsection{Challenges}
\label{subsec:challenges}

In our companion papers \citep{baryons_marginalization, Robust_marginalization} we show that the above architecture allows us to place a few percent constraints on the value of the cosmological parameters from almost all of the 13 different fields. However, we also encountered a series of obstacles and leave some work for future exploration. Here we outline what we believe are the most important challenges when doing parameter inference from CMD.

First, while our model is able to infer the value of the cosmological and astrophysical parameters with high accuracy for all the fields, the question remains of whether ours is the best model that can be constructed. It would be interesting to find models that perform better, i.e.~that can constrain the values of the parameters with higher accuracy. Knowing these constraints will allow us to better understand a deep and crucial question in cosmology: How much information do non-linear Gaussian fields contain? The codes and weights released in this work can be taken as a benchmark to improve upon.

Second, we found that for some fields, e.g.~gas temperature, if we train the model using IllustrisTNG maps and test it on SIMBA maps (and vice-versa), it is not able to recover the true value of the cosmological parameters. It is crucial for the model to be robust to the training data since models trained on simulated data may always face this problem, as simulations may never be perfect representations of reality. Thus, the second challenge is to find \textit{robust} models that can be trained on a given simulation dataset and will be able to perform inference when testing on a different dataset.

Third, in our companion paper \citep{baryons_marginalization} we found that when doing parameter inference over 2D maps with different fields as different channels (\textit{multifield}) the constraints on the parameters improve with respect to a single field. However, the improvement is relatively modest, i.e. we did not observe a major improvement. Thus, it will be interesting to find the minimum set of fields of a multifield that contain, e.g., 90\% and 95\% of the cosmological and possibly astrophysical information. In other words, if using the 13 different fields allows us to place a given constraint on the value of the cosmological and astrophysical parameters, what minimum subset of these fields enables us to get a fraction of those constraints? The idea behind this exercise is that those subsets of fields can be used to infer the value of the cosmological and astrophysical parameters, and the rest of the fields can be used to perform internal cross-validations. For instance, if the gas temperature field is not used to infer the parameter values, it still can be used when running simulations with the most likely parameter values to compare directly against observations, providing an additional internal check to verify the robustness of the model. 

Fourth, it would be very important to understand where the information from the different fields is coming from. In other words, what summary statistic, if any, are the CNNs using in order to constrain the value of the cosmological parameters? Are they focusing their attention into regions with large values of the considered field? Shedding light into this question will help to develop analytic methods to more robustly extract that information but also help us in understanding the process of non-linear gravitational evolution.

Of course, the above challenges apply to both 2D maps and 3D grids.

\subsection{Generative models}

While CMD data spans a wide range in the values of cosmological and astrophysical parameters, there may be some applications where more data is needed at points in parameter space not covered by CMD data. In this case, one can use techniques such as conditional GANs or conditional normalizing flows to generate new 2D maps or 3D grids conditioned on the values of the parameters \citep{Tamosiunas}. These emulators at the field level can be used in place of the more expensive simulations to carry out different tasks.

\subsection{N-body to hydrodynamic}

(Magneto-)hydrodynamic simulations are much more computationally expensive than gravity-only N-body simulations. At the time of writing this paper, running full hydrodynamic simulations over gigaparsec volumes with a reasonable resolution is computationally unfeasible. On the other hand, the N-body counterparts of these simulations are presently beginning to be feasible with advanced supercomputers. Thus, it may be desirable to \textit{paint} gas and star properties onto the dark matter field from N-body simulations (i.e. to transform dark matter into gas and star properties), as in \cite{Jay_2019, Leander_2020, Zhang_2019, Jacky_2019, Noah_2020, Harrington_2021, Yonseok_2019,Jay_2020}. Alternatively, the total matter field from the N-body simulations can be mapped to the total matter of the full hydrodynamic simulation. This will be necessary for creating weak lensing maps that incorporate astrophysics effects at the field level.

Since the CMD spans a large volume in parameter space, the above mapping(s) can be done conditionally on the values of the cosmological and astrophysical parameters. Furthermore, the mapping can be done for several fields at the same time in order to take into account all cross-correlations among the fields.

\subsection{Super-resolution}

CMD provides 3D grids at three different resolutions for 13 different fields. It would be very interesting to train models that can take as input the low-resolution map or grid from a given field and output a higher resolution version of it. We emphasize that the three different grids for each redshift and field provided by CMD arise from the same simulation, i.e.~the mass and spatial resolution of the underlying simulation is the same.

Ideally, one would like to run a low-resolution hydrodynamic simulation and use a model that can produce a higher resolution version of it. This will be extremely valuable for the astrophysics community as the computational cost quickly increases with mass and spatial resolution. While this task has been carried out for gravity-only simulations \citep{Doogesh_2020, Yin_2020a, Yueying_2021}, it still remains to be performed for cosmological hydrodynamic simulations. While the only difference between the three different CMD 3D grids is in the resolution of the mesh size rather than the resolution of the underlying simulation, developing super-resolution methods for fields from cosmological hydrodynamical simulations using CMD has the potential to contribute towards the development of such methods for simulations of different resolutions.

\subsection{Time evolution}

With enough disk space, it is possible to save as many snapshots of a simulation as desired. In practice, the number of snapshots generated is limited due to disk space constraints. It would be very valuable to have a model that given a set of snapshots at some redshifts could output snapshots at other redshifts. This will be valuable for understanding the time evolution of some phenomena and for constructing merger trees and lightcones from the simulations. \cite{Chen_2020} showed that this is possible for gravity-only simulations. CMD provides a rich dataset to train models to carry out this task for many different fields of hydrodynamic simulations.

\section{Summary}
\label{sec:summary}

In this paper we have introduced the CAMELS Multifield Dataset, CMD, a large cosmological and astrophysical dataset that contains hundreds of thousands of 2D maps and 3D grids of 13 different fields: 1) gas mass, 2) gas velocity, 3) gas temperature, 4) gas pressure, 5) gas metallicity, 6) neutral hydrogen mass, 7) electron density, 8) magnetic fields, 9) magnesium over iron ratio, 10) dark matter mass, 11) dark matter velocity, 12) stellar mass, and 13) total matter mass. CMD has been created from simulations of the CAMELS project \citep{CAMELS}, a collection of more than 4,000 gravity-only N-body simulations and state-of-the-art hydrodynamic simulations from thousands of different universes. Each 2D map and 3D grid is described by two cosmological and four astrophysical parameters (only in the case of the hydrodynamic simulations). 

We have described a few applications of CMD: 1) parameter inference, 2) generative models, 3) mapping N-body to hydrodynamic, 4) super-resolution, and 5) time evolution. In our companion papers \citep{baryons_marginalization, Robust_marginalization} we used CMD to show that neural networks can extract information from the field while marginalizing over astrophysical effects for all CMD fields. In this paper we have described in detail the architecture of our model together with the training procedure. 

We release all CMD data (over 70 Terabytes), together with the codes and network weights for the parameter inference task carried out in our companion papers \citep{baryons_marginalization, Robust_marginalization}. We provide further technical details on how to download, read, and manipulate the data in \url{https://camels-multifield-dataset.readthedocs.io}. We hope that CMD can become a standard dataset for machine learning applications in cosmology and astrophysics.

In the future, we will create maps for directly observable quantities from weak lensing, thermal and kinetic Sunyaev–Zeldovich effects, and X-ray and 21\,cm emission. Future public data releases from the CAMELS simulation suite will also include a database specifically targeted at multi-wavelength probes of circumgalactic medium profiles.

\section*{ACKNOWLEDGEMENTS}
FVN acknowledges funding from the WFIRST program through NNG26PJ30C and NNN12AA01C. DAA was supported in part by NSF grants AST-2009687 and AST-2108944. The work of DNS, SG, DAA, LT, YL, AN, SH, and BDW has been supported by the Simons Foundation. The work of PVD is supported by CIDEGENT/2018/019, CPI-21-108. The 2D maps and 3D grids have been created using the Pylians3 libraries \url{https://github.com/franciscovillaescusa/Pylians3} and voxelize \url{https://github.com/leanderthiele/voxelize}. CMD has been created using the Rusty cluster of the Flatiron Institute. Details on the CAMELS simulations can be found in \url{https://www.camel-simulations.org}.

\bibliography{references}{}
\bibliographystyle{aasjournal}

\end{document}